\documentclass[letterpaper]{article} 

\usepackage{preamble/ijcai24}

\usepackage{times}
\usepackage{soul}
\usepackage{url}
\usepackage[hidelinks]{hyperref}
\usepackage[utf8]{inputenc}
\usepackage[small]{caption}
\usepackage{graphicx}
\usepackage{amsmath}
\usepackage{amsthm}
\usepackage{booktabs}
\usepackage[switch]{lineno}

\ifdefined\SUBMISSION
    \linenumbers
\fi

\usepackage{svg}
\usepackage{xcolor}
\usepackage{graphicx}
\graphicspath{{figures/}}
\usepackage{amsmath}
\usepackage{amsfonts}
\usepackage{cleveref}
\usepackage{svg}
\usepackage{caption}
\usepackage{subcaption}
\usepackage{adjustbox}
\usepackage{longtable,tabularx,booktabs}
\usepackage[flushleft]{threeparttable}
\usepackage{multirow}
\usepackage{lipsum}
\usepackage{makecell}
\usepackage{mdframed}
\usepackage{framed}
\usepackage{comment}
\usepackage{dsfont} 
\usepackage{preamble/vector} 
\usepackage{mathtools}
\usepackage[title]{appendix}
\usepackage[bottom]{footmisc} 
\usepackage{colortbl}
\makeatletter
\@namedef{ver@fixltx2e.sty}{}
\makeatother
\usepackage{dblfloatfix}
\usepackage{floatrow}
\newfloatcommand{capbtabbox}{table}[][\FBwidth] 

\newcommand{\citet}[1]{\citeauthor{#1}~\shortcite{#1}}

\DeclareFontFamily{U}{mathx}{\hyphenchar\font45}
\DeclareFontShape{U}{mathx}{m}{n}{
      <5> <6> <7> <8> <9> <10>
      <10.95> <12> <14.4> <17.28> <20.74> <24.88>
      mathx10
      }{}
\DeclareSymbolFont{mathx}{U}{mathx}{m}{n}
\DeclareFontSubstitution{U}{mathx}{m}{n}

\DeclareMathSymbol{\betterbigvee}         {1}{mathx}{"9A}
\DeclareMathSymbol{\betterbigwedge}       {1}{mathx}{"99}

\usepackage{algorithmicx}
\usepackage{algorithm}
\usepackage{algpseudocode}

\algdef{SE}[FOR]{ParallelFor}{EndParallelFor}[1]{\textbf{parallel} \algorithmicfor\ #1}{\algorithmicend}%
\algdef{SE}[FOR]{For}{EndFor}[1]{\algorithmicfor\ #1}{\algorithmicend}%
\algdef{SE}[IF]{If}{EndIf}[1]{\algorithmicif\ #1}{\algorithmicend}%
\algdef{C}[IF]{IF}{ElsIf}[1]{\algorithmicelse\ \algorithmicif\ #1}%
\algdef{Ce}[ELSE]{IF}{Else}{EndIf}{\algorithmicelse}%
\algdef{SE}[FUNCTION]{Function}{EndFunction}%
   [2]{\algorithmicfunction\ \textproc{#1}\ifthenelse{\equal{#2}{}}{}{(#2)}}%
   {\algorithmicend}%
\algtext*{EndFunction} 
\algtext*{EndFor} 
\algtext*{EndParallelFor} 
\algtext*{EndIf} 

\definecolor{commentgray}{rgb}{0, 0, 0}
\newcommand{\GrayComment}[1]{{\hfill{\color{commentgray}$\triangleright$ #1}}}

\algrenewcommand\alglinenumber[1]{\color{commentgray}\tiny #1}

\title{ConstrainedZero: Chance-Constrained POMDP Planning using Learned\\Probabilistic Failure Surrogates and Adaptive Safety Constraints}

\ifdefined\SUBMISSION
    \author{
        Anonymous Author(s)\\
        \ Anonymous Author(s)
        \affiliations
        Affiliation(s)\\
        \ Affiliation(s)
        \emails
        email@example.com
    }
\else
    \author{
        Robert J. Moss$^1${\rm ,}\enspace
        Arec Jamgochian$^1${\rm ,}\enspace
        Johannes Fischer$^{1,2}${\rm ,}\\
        Anthony Corso$^1${\rm ,}\And
        Mykel J. Kochenderfer$^1$
        \affiliations
        $^1$Stanford University, Stanford, CA\\
        $^2$Karlsruhe Institute of Technology (KIT), Karlsruhe, Germany
        \emails
        \{mossr,\,arec,\,acorso,\,mykel\}@stanford.edu \quad johannes.fischer@kit.edu
    }
\fi

\usepackage{tikz}
\usepackage{pgfplots}
\pgfplotsset{compat=newest}

\usetikzlibrary{calc}
\usetikzlibrary{shapes.geometric}
\usetikzlibrary{external}
\usetikzlibrary{patterns}
\usetikzlibrary{shapes,arrows,fit}
\usetikzlibrary{positioning}
\usetikzlibrary{arrows.meta, calc, shapes}
\usetikzlibrary{graphs}
\usetikzlibrary{decorations.pathmorphing}
\usetikzlibrary{decorations.pathreplacing}
\usetikzlibrary{backgrounds}
\usetikzlibrary{shadows}

\definecolor{blues1}{RGB}{198, 219, 239}
\definecolor{blues2}{RGB}{158, 202, 225}
\definecolor{blues3}{RGB}{107, 174, 214}
\definecolor{blues4}{RGB}{49, 130, 189}
\definecolor{blues5}{RGB}{8, 81, 156}

\definecolor{grays1}{RGB}{219, 219, 219}
\definecolor{grays2}{RGB}{202, 202, 202}
\definecolor{grays3}{RGB}{174, 174, 174}
\definecolor{grays4}{RGB}{130, 130, 130}
\definecolor{grays5}{RGB}{81, 81, 81}

\usepackage{pagecolor}

\ifdefined\INVERTED

\else

\fi

\usetikzlibrary{arrows.meta}
\usetikzlibrary{backgrounds}
\usepgfplotslibrary{patchplots}
\usepgfplotslibrary{fillbetween}
\pgfplotsset{%
    layers/standard/.define layer set={%
        background,axis background,axis grid,axis ticks,axis lines,axis tick labels,pre main,main,axis descriptions,axis foreground%
    }{
        grid style={/pgfplots/on layer=axis grid},%
        tick style={/pgfplots/on layer=axis ticks},%
        axis line style={/pgfplots/on layer=axis lines},%
        label style={/pgfplots/on layer=axis descriptions},%
        legend style={/pgfplots/on layer=axis descriptions},%
        title style={/pgfplots/on layer=axis descriptions},%
        colorbar style={/pgfplots/on layer=axis descriptions},%
        ticklabel style={/pgfplots/on layer=axis tick labels},%
        axis background@ style={/pgfplots/on layer=axis background},%
        3d box foreground style={/pgfplots/on layer=axis foreground},%
    },
}

\pgfplotsset{
colormap={plots1}{rgb(0.00000000)=(0.26700400,0.00487400,0.32941500)
rgb(0.00392157)=(0.26851000,0.00960500,0.33542700)
rgb(0.00784314)=(0.26994400,0.01462500,0.34137900)
rgb(0.01176471)=(0.27130500,0.01994200,0.34726900)
rgb(0.01568627)=(0.27259400,0.02556300,0.35309300)
rgb(0.01960784)=(0.27380900,0.03149700,0.35885300)
rgb(0.02352941)=(0.27495200,0.03775200,0.36454300)
rgb(0.02745098)=(0.27602200,0.04416700,0.37016400)
rgb(0.03137255)=(0.27701800,0.05034400,0.37571500)
rgb(0.03529412)=(0.27794100,0.05632400,0.38119100)
rgb(0.03921569)=(0.27879100,0.06214500,0.38659200)
rgb(0.04313725)=(0.27956600,0.06783600,0.39191700)
rgb(0.04705882)=(0.28026700,0.07341700,0.39716300)
rgb(0.05098039)=(0.28089400,0.07890700,0.40232900)
rgb(0.05490196)=(0.28144600,0.08432000,0.40741400)
rgb(0.05882353)=(0.28192400,0.08966600,0.41241500)
rgb(0.06274510)=(0.28232700,0.09495500,0.41733100)
rgb(0.06666667)=(0.28265600,0.10019600,0.42216000)
rgb(0.07058824)=(0.28291000,0.10539300,0.42690200)
rgb(0.07450980)=(0.28309100,0.11055300,0.43155400)
rgb(0.07843137)=(0.28319700,0.11568000,0.43611500)
rgb(0.08235294)=(0.28322900,0.12077700,0.44058400)
rgb(0.08627451)=(0.28318700,0.12584800,0.44496000)
rgb(0.09019608)=(0.28307200,0.13089500,0.44924100)
rgb(0.09411765)=(0.28288400,0.13592000,0.45342700)
rgb(0.09803922)=(0.28262300,0.14092600,0.45751700)
rgb(0.10196078)=(0.28229000,0.14591200,0.46151000)
rgb(0.10588235)=(0.28188700,0.15088100,0.46540500)
rgb(0.10980392)=(0.28141200,0.15583400,0.46920100)
rgb(0.11372549)=(0.28086800,0.16077100,0.47289900)
rgb(0.11764706)=(0.28025500,0.16569300,0.47649800)
rgb(0.12156863)=(0.27957400,0.17059900,0.47999700)
rgb(0.12549020)=(0.27882600,0.17549000,0.48339700)
rgb(0.12941176)=(0.27801200,0.18036700,0.48669700)
rgb(0.13333333)=(0.27713400,0.18522800,0.48989800)
rgb(0.13725490)=(0.27619400,0.19007400,0.49300100)
rgb(0.14117647)=(0.27519100,0.19490500,0.49600500)
rgb(0.14509804)=(0.27412800,0.19972100,0.49891100)
rgb(0.14901961)=(0.27300600,0.20452000,0.50172100)
rgb(0.15294118)=(0.27182800,0.20930300,0.50443400)
rgb(0.15686275)=(0.27059500,0.21406900,0.50705200)
rgb(0.16078431)=(0.26930800,0.21881800,0.50957700)
rgb(0.16470588)=(0.26796800,0.22354900,0.51200800)
rgb(0.16862745)=(0.26658000,0.22826200,0.51434900)
rgb(0.17254902)=(0.26514500,0.23295600,0.51659900)
rgb(0.17647059)=(0.26366300,0.23763100,0.51876200)
rgb(0.18039216)=(0.26213800,0.24228600,0.52083700)
rgb(0.18431373)=(0.26057100,0.24692200,0.52282800)
rgb(0.18823529)=(0.25896500,0.25153700,0.52473600)
rgb(0.19215686)=(0.25732200,0.25613000,0.52656300)
rgb(0.19607843)=(0.25564500,0.26070300,0.52831200)
rgb(0.20000000)=(0.25393500,0.26525400,0.52998300)
rgb(0.20392157)=(0.25219400,0.26978300,0.53157900)
rgb(0.20784314)=(0.25042500,0.27429000,0.53310300)
rgb(0.21176471)=(0.24862900,0.27877500,0.53455600)
rgb(0.21568627)=(0.24681100,0.28323700,0.53594100)
rgb(0.21960784)=(0.24497200,0.28767500,0.53726000)
rgb(0.22352941)=(0.24311300,0.29209200,0.53851600)
rgb(0.22745098)=(0.24123700,0.29648500,0.53970900)
rgb(0.23137255)=(0.23934600,0.30085500,0.54084400)
rgb(0.23529412)=(0.23744100,0.30520200,0.54192100)
rgb(0.23921569)=(0.23552600,0.30952700,0.54294400)
rgb(0.24313725)=(0.23360300,0.31382800,0.54391400)
rgb(0.24705882)=(0.23167400,0.31810600,0.54483400)
rgb(0.25098039)=(0.22973900,0.32236100,0.54570600)
rgb(0.25490196)=(0.22780200,0.32659400,0.54653200)
rgb(0.25882353)=(0.22586300,0.33080500,0.54731400)
rgb(0.26274510)=(0.22392500,0.33499400,0.54805300)
rgb(0.26666667)=(0.22198900,0.33916100,0.54875200)
rgb(0.27058824)=(0.22005700,0.34330700,0.54941300)
rgb(0.27450980)=(0.21813000,0.34743200,0.55003800)
rgb(0.27843137)=(0.21621000,0.35153500,0.55062700)
rgb(0.28235294)=(0.21429800,0.35561900,0.55118400)
rgb(0.28627451)=(0.21239500,0.35968300,0.55171000)
rgb(0.29019608)=(0.21050300,0.36372700,0.55220600)
rgb(0.29411765)=(0.20862300,0.36775200,0.55267500)
rgb(0.29803922)=(0.20675600,0.37175800,0.55311700)
rgb(0.30196078)=(0.20490300,0.37574600,0.55353300)
rgb(0.30588235)=(0.20306300,0.37971600,0.55392500)
rgb(0.30980392)=(0.20123900,0.38367000,0.55429400)
rgb(0.31372549)=(0.19943000,0.38760700,0.55464200)
rgb(0.31764706)=(0.19763600,0.39152800,0.55496900)
rgb(0.32156863)=(0.19586000,0.39543300,0.55527600)
rgb(0.32549020)=(0.19410000,0.39932300,0.55556500)
rgb(0.32941176)=(0.19235700,0.40319900,0.55583600)
rgb(0.33333333)=(0.19063100,0.40706100,0.55608900)
rgb(0.33725490)=(0.18892300,0.41091000,0.55632600)
rgb(0.34117647)=(0.18723100,0.41474600,0.55654700)
rgb(0.34509804)=(0.18555600,0.41857000,0.55675300)
rgb(0.34901961)=(0.18389800,0.42238300,0.55694400)
rgb(0.35294118)=(0.18225600,0.42618400,0.55712000)
rgb(0.35686275)=(0.18062900,0.42997500,0.55728200)
rgb(0.36078431)=(0.17901900,0.43375600,0.55743000)
rgb(0.36470588)=(0.17742300,0.43752700,0.55756500)
rgb(0.36862745)=(0.17584100,0.44129000,0.55768500)
rgb(0.37254902)=(0.17427400,0.44504400,0.55779200)
rgb(0.37647059)=(0.17271900,0.44879100,0.55788500)
rgb(0.38039216)=(0.17117600,0.45253000,0.55796500)
rgb(0.38431373)=(0.16964600,0.45626200,0.55803000)
rgb(0.38823529)=(0.16812600,0.45998800,0.55808200)
rgb(0.39215686)=(0.16661700,0.46370800,0.55811900)
rgb(0.39607843)=(0.16511700,0.46742300,0.55814100)
rgb(0.40000000)=(0.16362500,0.47113300,0.55814800)
rgb(0.40392157)=(0.16214200,0.47483800,0.55814000)
rgb(0.40784314)=(0.16066500,0.47854000,0.55811500)
rgb(0.41176471)=(0.15919400,0.48223700,0.55807300)
rgb(0.41568627)=(0.15772900,0.48593200,0.55801300)
rgb(0.41960784)=(0.15627000,0.48962400,0.55793600)
rgb(0.42352941)=(0.15481500,0.49331300,0.55784000)
rgb(0.42745098)=(0.15336400,0.49700000,0.55772400)
rgb(0.43137255)=(0.15191800,0.50068500,0.55758700)
rgb(0.43529412)=(0.15047600,0.50436900,0.55743000)
rgb(0.43921569)=(0.14903900,0.50805100,0.55725000)
rgb(0.44313725)=(0.14760700,0.51173300,0.55704900)
rgb(0.44705882)=(0.14618000,0.51541300,0.55682300)
rgb(0.45098039)=(0.14475900,0.51909300,0.55657200)
rgb(0.45490196)=(0.14334300,0.52277300,0.55629500)
rgb(0.45882353)=(0.14193500,0.52645300,0.55599100)
rgb(0.46274510)=(0.14053600,0.53013200,0.55565900)
rgb(0.46666667)=(0.13914700,0.53381200,0.55529800)
rgb(0.47058824)=(0.13777000,0.53749200,0.55490600)
rgb(0.47450980)=(0.13640800,0.54117300,0.55448300)
rgb(0.47843137)=(0.13506600,0.54485300,0.55402900)
rgb(0.48235294)=(0.13374300,0.54853500,0.55354100)
rgb(0.48627451)=(0.13244400,0.55221600,0.55301800)
rgb(0.49019608)=(0.13117200,0.55589900,0.55245900)
rgb(0.49411765)=(0.12993300,0.55958200,0.55186400)
rgb(0.49803922)=(0.12872900,0.56326500,0.55122900)
rgb(0.50196078)=(0.12756800,0.56694900,0.55055600)
rgb(0.50588235)=(0.12645300,0.57063300,0.54984100)
rgb(0.50980392)=(0.12539400,0.57431800,0.54908600)
rgb(0.51372549)=(0.12439500,0.57800200,0.54828700)
rgb(0.51764706)=(0.12346300,0.58168700,0.54744500)
rgb(0.52156863)=(0.12260600,0.58537100,0.54655700)
rgb(0.52549020)=(0.12183100,0.58905500,0.54562300)
rgb(0.52941176)=(0.12114800,0.59273900,0.54464100)
rgb(0.53333333)=(0.12056500,0.59642200,0.54361100)
rgb(0.53725490)=(0.12009200,0.60010400,0.54253000)
rgb(0.54117647)=(0.11973800,0.60378500,0.54140000)
rgb(0.54509804)=(0.11951200,0.60746400,0.54021800)
rgb(0.54901961)=(0.11942300,0.61114100,0.53898200)
rgb(0.55294118)=(0.11948300,0.61481700,0.53769200)
rgb(0.55686275)=(0.11969900,0.61849000,0.53634700)
rgb(0.56078431)=(0.12008100,0.62216100,0.53494600)
rgb(0.56470588)=(0.12063800,0.62582800,0.53348800)
rgb(0.56862745)=(0.12138000,0.62949200,0.53197300)
rgb(0.57254902)=(0.12231200,0.63315300,0.53039800)
rgb(0.57647059)=(0.12344400,0.63680900,0.52876300)
rgb(0.58039216)=(0.12478000,0.64046100,0.52706800)
rgb(0.58431373)=(0.12632600,0.64410700,0.52531100)
rgb(0.58823529)=(0.12808700,0.64774900,0.52349100)
rgb(0.59215686)=(0.13006700,0.65138400,0.52160800)
rgb(0.59607843)=(0.13226800,0.65501400,0.51966100)
rgb(0.60000000)=(0.13469200,0.65863600,0.51764900)
rgb(0.60392157)=(0.13733900,0.66225200,0.51557100)
rgb(0.60784314)=(0.14021000,0.66585900,0.51342700)
rgb(0.61176471)=(0.14330300,0.66945900,0.51121500)
rgb(0.61568627)=(0.14661600,0.67305000,0.50893600)
rgb(0.61960784)=(0.15014800,0.67663100,0.50658900)
rgb(0.62352941)=(0.15389400,0.68020300,0.50417200)
rgb(0.62745098)=(0.15785100,0.68376500,0.50168600)
rgb(0.63137255)=(0.16201600,0.68731600,0.49912900)
rgb(0.63529412)=(0.16638300,0.69085600,0.49650200)
rgb(0.63921569)=(0.17094800,0.69438400,0.49380300)
rgb(0.64313725)=(0.17570700,0.69790000,0.49103300)
rgb(0.64705882)=(0.18065300,0.70140200,0.48818900)
rgb(0.65098039)=(0.18578300,0.70489100,0.48527300)
rgb(0.65490196)=(0.19109000,0.70836600,0.48228400)
rgb(0.65882353)=(0.19657100,0.71182700,0.47922100)
rgb(0.66274510)=(0.20221900,0.71527200,0.47608400)
rgb(0.66666667)=(0.20803000,0.71870100,0.47287300)
rgb(0.67058824)=(0.21400000,0.72211400,0.46958800)
rgb(0.67450980)=(0.22012400,0.72550900,0.46622600)
rgb(0.67843137)=(0.22639700,0.72888800,0.46278900)
rgb(0.68235294)=(0.23281500,0.73224700,0.45927700)
rgb(0.68627451)=(0.23937400,0.73558800,0.45568800)
rgb(0.69019608)=(0.24607000,0.73891000,0.45202400)
rgb(0.69411765)=(0.25289900,0.74221100,0.44828400)
rgb(0.69803922)=(0.25985700,0.74549200,0.44446700)
rgb(0.70196078)=(0.26694100,0.74875100,0.44057300)
rgb(0.70588235)=(0.27414900,0.75198800,0.43660100)
rgb(0.70980392)=(0.28147700,0.75520300,0.43255200)
rgb(0.71372549)=(0.28892100,0.75839400,0.42842600)
rgb(0.71764706)=(0.29647900,0.76156100,0.42422300)
rgb(0.72156863)=(0.30414800,0.76470400,0.41994300)
rgb(0.72549020)=(0.31192500,0.76782200,0.41558600)
rgb(0.72941176)=(0.31980900,0.77091400,0.41115200)
rgb(0.73333333)=(0.32779600,0.77398000,0.40664000)
rgb(0.73725490)=(0.33588500,0.77701800,0.40204900)
rgb(0.74117647)=(0.34407400,0.78002900,0.39738100)
rgb(0.74509804)=(0.35236000,0.78301100,0.39263600)
rgb(0.74901961)=(0.36074100,0.78596400,0.38781400)
rgb(0.75294118)=(0.36921400,0.78888800,0.38291400)
rgb(0.75686275)=(0.37777900,0.79178100,0.37793900)
rgb(0.76078431)=(0.38643300,0.79464400,0.37288600)
rgb(0.76470588)=(0.39517400,0.79747500,0.36775700)
rgb(0.76862745)=(0.40400100,0.80027500,0.36255200)
rgb(0.77254902)=(0.41291300,0.80304100,0.35726900)
rgb(0.77647059)=(0.42190800,0.80577400,0.35191000)
rgb(0.78039216)=(0.43098300,0.80847300,0.34647600)
rgb(0.78431373)=(0.44013700,0.81113800,0.34096700)
rgb(0.78823529)=(0.44936800,0.81376800,0.33538400)
rgb(0.79215686)=(0.45867400,0.81636300,0.32972700)
rgb(0.79607843)=(0.46805300,0.81892100,0.32399800)
rgb(0.80000000)=(0.47750400,0.82144400,0.31819500)
rgb(0.80392157)=(0.48702600,0.82392900,0.31232100)
rgb(0.80784314)=(0.49661500,0.82637600,0.30637700)
rgb(0.81176471)=(0.50627100,0.82878600,0.30036200)
rgb(0.81568627)=(0.51599200,0.83115800,0.29427900)
rgb(0.81960784)=(0.52577600,0.83349100,0.28812700)
rgb(0.82352941)=(0.53562100,0.83578500,0.28190800)
rgb(0.82745098)=(0.54552400,0.83803900,0.27562600)
rgb(0.83137255)=(0.55548400,0.84025400,0.26928100)
rgb(0.83529412)=(0.56549800,0.84243000,0.26287700)
rgb(0.83921569)=(0.57556300,0.84456600,0.25641500)
rgb(0.84313725)=(0.58567800,0.84666100,0.24989700)
rgb(0.84705882)=(0.59583900,0.84871700,0.24332900)
rgb(0.85098039)=(0.60604500,0.85073300,0.23671200)
rgb(0.85490196)=(0.61629300,0.85270900,0.23005200)
rgb(0.85882353)=(0.62657900,0.85464500,0.22335300)
rgb(0.86274510)=(0.63690200,0.85654200,0.21662000)
rgb(0.86666667)=(0.64725700,0.85840000,0.20986100)
rgb(0.87058824)=(0.65764200,0.86021900,0.20308200)
rgb(0.87450980)=(0.66805400,0.86199900,0.19629300)
rgb(0.87843137)=(0.67848900,0.86374200,0.18950300)
rgb(0.88235294)=(0.68894400,0.86544800,0.18272500)
rgb(0.88627451)=(0.69941500,0.86711700,0.17597100)
rgb(0.89019608)=(0.70989800,0.86875100,0.16925700)
rgb(0.89411765)=(0.72039100,0.87035000,0.16260300)
rgb(0.89803922)=(0.73088900,0.87191600,0.15602900)
rgb(0.90196078)=(0.74138800,0.87344900,0.14956100)
rgb(0.90588235)=(0.75188400,0.87495100,0.14322800)
rgb(0.90980392)=(0.76237300,0.87642400,0.13706400)
rgb(0.91372549)=(0.77285200,0.87786800,0.13110900)
rgb(0.91764706)=(0.78331500,0.87928500,0.12540500)
rgb(0.92156863)=(0.79376000,0.88067800,0.12000500)
rgb(0.92549020)=(0.80418200,0.88204600,0.11496500)
rgb(0.92941176)=(0.81457600,0.88339300,0.11034700)
rgb(0.93333333)=(0.82494000,0.88472000,0.10621700)
rgb(0.93725490)=(0.83527000,0.88602900,0.10264600)
rgb(0.94117647)=(0.84556100,0.88732200,0.09970200)
rgb(0.94509804)=(0.85581000,0.88860100,0.09745200)
rgb(0.94901961)=(0.86601300,0.88986800,0.09595300)
rgb(0.95294118)=(0.87616800,0.89112500,0.09525000)
rgb(0.95686275)=(0.88627100,0.89237400,0.09537400)
rgb(0.96078431)=(0.89632000,0.89361600,0.09633500)
rgb(0.96470588)=(0.90631100,0.89485500,0.09812500)
rgb(0.96862745)=(0.91624200,0.89609100,0.10071700)
rgb(0.97254902)=(0.92610600,0.89733000,0.10407100)
rgb(0.97647059)=(0.93590400,0.89857000,0.10813100)
rgb(0.98039216)=(0.94563600,0.89981500,0.11283800)
rgb(0.98431373)=(0.95530000,0.90106500,0.11812800)
rgb(0.98823529)=(0.96489400,0.90232300,0.12394100)
rgb(0.99215686)=(0.97441700,0.90359000,0.13021500)
rgb(0.99607843)=(0.98386800,0.90486700,0.13689700)
rgb(1.00000000)=(0.99324800,0.90615700,0.14393600)},
}

\definecolor{timingcolor}{rgb}{0.35, 0.35, 0.35}

\DeclareMathOperator*{\argmax}{arg\,max}

\DeclarePairedDelimiter{\norm}{\lVert}{\rVert}

\newcommand*{\defeq}{\stackrel{\text{def}}{=}}

\newcommand*\diff{\mathop{}\!\mathrm{d}}
\newcommand{\bpi}{\boldsymbol\pi}

\definecolor{resultcolor}{HTML}{000000}
\newsavebox\CBox
\def\mathBF#1{\sbox\CBox{$#1$}\resizebox{\wd\CBox}{\ht\CBox}{{\color{resultcolor}$\mathbf{#1}$}}}

\usepackage{siunitx}
\makeatletter
\let\amsmath@bigmidtmp\bigm

\newcommand{\bigmid}[1]{%
  \ifcsname fenced@\string#1\endcsname
    \expandafter\@firstoftwo
  \else
    \expandafter\@secondoftwo
  \fi
  {\expandafter\amsmath@bigmidtmp\csname fenced@\string#1\endcsname}%
  {\amsmath@bigmidtmp#1}%
}
\newcommand{\DeclareFence}[2]{\@namedef{fenced@\string#1}{#2}}
\makeatother

\DeclareFence{\mid}{|}

\definecolor{newcolor}{HTML}{C70039} 

\newcommand{\new}[1]{{\color{newcolor}#1}}
\newcommand{\bpm}{\boldsymbol\pm}
\definecolor{tablecolorgood}{HTML}{007662}
\definecolor{tablecolorbad}{HTML}{8c1515}

\begin{document}

\ifdefined\APPENDIXONLY
    \setcounter{equation}{21} 


\section*{Appendix}
The following sections detail additional empirical analysis, connections between our simplified application of adaptive conformal inference (ACI) and how it is equivalent to quantile coverage, and information regarding the hyperparameters and neural network architectures.

\appendix

\subsection{Empirical analysis of ACI step size \texorpdfstring{$\eta$}{η} on training}
To test the sensitivity of ConstrainedZero to the ACI step size $\eta$, we swept values of $\eta$ for the LightDark CC-POMDP (\cref{fig:eta_sweep}).
The step size controls the reactivity of the updates in the adaptation step of $\Delta$-MCTS.
\Cref{fig:eta_returns} shows that with a larger step size, the reactivity of the threshold update in \ifdefined\APPENDIXONLY eq.~(18) \else\cref{eq:aci_update}\fi opens the safety threshold faster, resulting in more risky behavior at the expense of higher returns.
Due to its stability, a step size of $\eta = \num{1e-5}$ was chosen for the final results in \ifdefined\APPENDIXONLY table 1\else\cref{tab:results}\fi.
Future research may focus on methods for adapting the step size online during planning, such as the parameter-free method AgACI from \citet{zaffran2022adaptive}.

\begin{figure}[ht]
    \ifdefined\APPENDIXONLY
        \phantom{---}

        \phantom{---}
    \fi
    \centering
    \begin{subfigure}[c]{\textwidth}
        \centering
        \resizebox{\ifdefined\APPENDIXONLY\else0.8\fi\textwidth}{!}{%
            \includegraphics{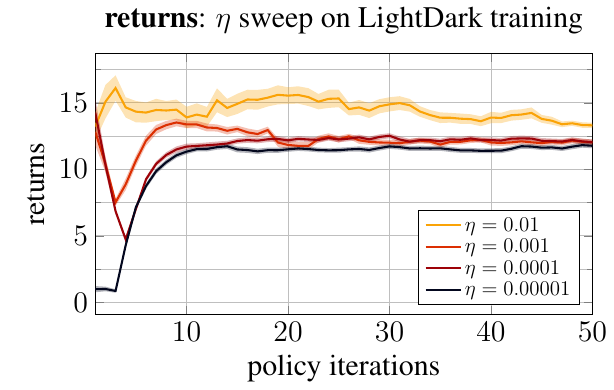}
        }
        \caption{Returns from LightDark policy iteration when sweeping $\eta$.}\label{fig:eta_returns}
    \end{subfigure}
    \phantom{---}

    \phantom{---}

    \phantom{---}
    \begin{subfigure}[c]{\textwidth}
        \centering
        \resizebox{\ifdefined\APPENDIXONLY\else0.8\fi\textwidth}{!}{%
            \includegraphics{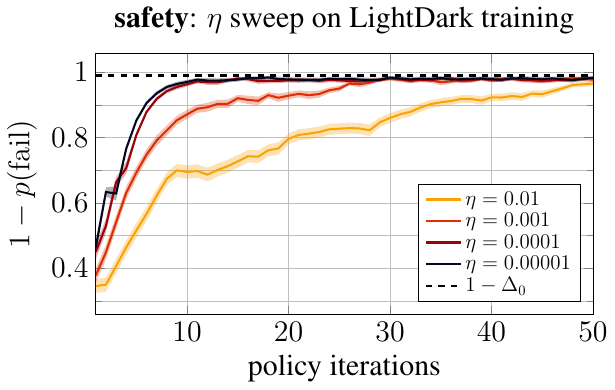}
        }
        \caption{Safety from LightDark policy iteration when sweeping $\eta$.}\label{fig:eta_safety}
    \end{subfigure}
    \caption{Empirical sensitivity analysis of the ACI step size $\eta$ on offline policy iteration (i.e., training) for the LightDark CC-POMDP. The larger the $\eta$, the more reactive the ACI update will be.}\label{fig:eta_sweep}
\end{figure}

\subsection{Connection to ACI quantiles}
In adaptive conformal inference (ACI), the algorithm provides coverage based on quantiles of an estimated value from some data distribution \cite{gibbs2021adaptive}.
In our work, we simplify the ACI formulation.
To connect to the original ACI formulation, let $F(b,a)$ be the estimated failure probability ($F$-value) and $\hat{Q}(\cdot)$ be the quantile function where the $\Delta$-quantile is defined on the range $[0,1]$.\footnote{The quantile function is typically defined over an input set $Y$. But in our case, it is defined over an input range $[0,1]$ to assess probabilities in a standardized way.}
Let the set of failure probabilities associated to a belief-state node $b$ with children $A(b)$ be
\begin{equation}
    \mathbf{f} = \big\{F(b,a') : a' \in A(b)\big\}.
\end{equation}
In this formulation, we would want to update $\Delta(b)$ based on the error as indication of miscoverage of the most recent $F$-value $F(b,a)$:
\begin{equation}
    \text{err} = \begin{cases}
        1 & \text{if } F(b,a) \not\in \hat{C}_{\Delta} \\
        0 & \text{otherwise}
    \end{cases}
\end{equation}
where $\hat{C}_{\Delta} := \big\{ p_i : p_i \le \hat{Q}(\Delta(b)),\ p_i \in \mathbf{f} \big\}$ is the covered set. Here we use $\Delta(b)$ instead of $1 - \Delta(b)$ from \citet{gibbs2021adaptive} because we want the lower quantile (i.e., we want to be below the failure probability threshold).
This is equivalent to our simplification where the error is defined as the miscoverage of the $F$-value by the $\Delta(b)$ estimate
\begin{equation}
    \text{err} = \mathds{1}\{F(b,a) > \Delta(b)\}
\end{equation}
and using the same update as $\Delta$-MCTS of
\begin{equation}
    \Delta(b) = \Delta(b) + \eta(\text{err} - \Delta_0).
\end{equation}
Note that we reformulate the update function for our setting.
In the original ACI formulation, the update is done according to $\Delta(b) = \Delta(b) + \eta(\Delta_0 - \text{err})$.
In the original version, if the error is zero (indicating coverage), then $\Delta(b)$ is increased by $\eta\Delta_0$ (becoming tighter, noting this assumes the coverage is based on the upper quantile).
If the error is one (indicating miscoverage), then $\Delta(b)$ is decreased by $\eta(1 - \Delta_0)$ (widening the coverage).
In our version, we reverse the update to operate on the lower quantile, keeping the reactivity of the algorithm during miscoverage events (thus, increasing by $\eta(1-\Delta_0)$ in this case, as described in \ifdefined\APPENDIXONLY eq.~(19)\else\cref{eq:err}\fi).

\subsection{Hyperparameters}
The parameters used for ConstrainedZero and $\Delta$-MCTS are included along with the code for all experiments and CC-POMDP environments in the BetaZero.jl Julia package.\footnote{\url{https://github.com/sisl/BetaZero.jl/tree/safety}}

\paragraph{Discussion of weight parameter \texorpdfstring{$\delta$}{δ}.}
When computing the total failure probability given the immediate probability $p$ and the future probability $p'$, we apply a weight $\delta$, or discount, to the final estimate (\ifdefined\APPENDIXONLY eq.~(14)\else\cref{eq:p_fail}\fi).
It is well known that the discount can be interpreted as the $(1 - \delta)$ probability of termination on the next step \cite{littman1994markov,sutton2018reinforcement}.

\subsection{Neural network architecture}
The neural network used for each CC-POMDP is a simple fully-connected feedforward network shown in \cref{fig:nn_arch} and based on the network used by BetaZero \cite{moss2023betazero}.
For the LightDark problem, an input size of $m=2$ is used for the mean and standard deviation of the particle filter belief over the $y$-location state values, with an internal network depth of $d=2$ and width of $k=64$.
For the CAS problem, an input size of $m=20$ is used for the mean and covariance of the unscented Kalman filter belief over the state variables of $h_\text{rel}$, $\dot{h}_\text{rel}$, $a_\text{prev}$, and $\tau$, with a network depth of $d=2$ and width of $k=64$.
Finally, the spillpoint CCS problem uses an input size of $m=510$ (mean and standard deviation for the top-surface grid points, top-surface heights, porosity, injection locations, and injection depth), with a network depth of $d=4$ and width of $k=128$.

Following \citet{moss2023betazero}, the value head of the network is trained on the normalized returns so that they lie in the range of $[-1, 1]$.
An output denormalization layer is appended to the value head so that the predictions are properly scaled (done entirely internal to the network).
Value normalization is useful for training stability so that the training targets have zero mean and unit variance \cite{lecun2002efficient}.

\begin{figure}[ht]
    \centering
        \resizebox{\textwidth}{!}{%
            \includegraphics{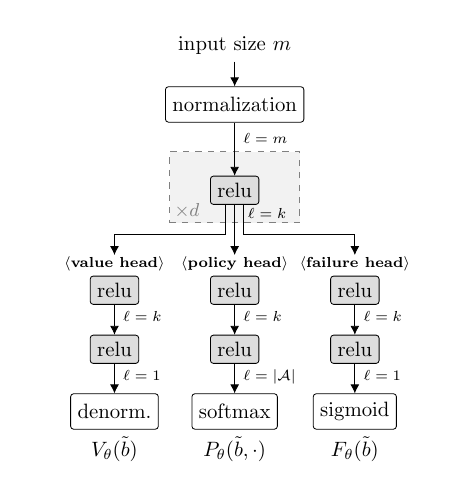}
        }
    \caption{Simple network architecture used for each CC-POMDP.}
    \label{fig:nn_arch}
\end{figure}

    \bibliographystyle{preamble/named}
    \bibliography{references}
\else
    \maketitle

    \begin{abstract}
To plan safely in uncertain environments, agents must balance utility with safety constraints.
Safe planning problems can be modeled as a chance-constrained partially observable Markov decision process (CC-POMDP) and solutions often use expensive rollouts or heuristics to estimate the optimal value and action-selection policy.
This work introduces the \textit{ConstrainedZero} policy iteration algorithm that solves CC-POMDPs in belief space by learning neural network approximations of the optimal value and policy with an additional network head that estimates the failure probability given a belief.
This failure probability guides safe action selection during online Monte Carlo tree search (MCTS).
To avoid overemphasizing search based on the failure estimates, we introduce $\Delta\text{-MCTS}$, which uses adaptive conformal inference to update the failure threshold during planning.
The approach is tested on a safety-critical POMDP benchmark, an aircraft collision avoidance system, and the sustainability problem of safe CO$_2$ storage.
Results show that by separating safety constraints from the objective we can achieve a target level of safety without optimizing the balance between rewards and costs.
\end{abstract}

\begin{figure}[t]
    \centering
    \resizebox{\linewidth}{!}{
        \includegraphics{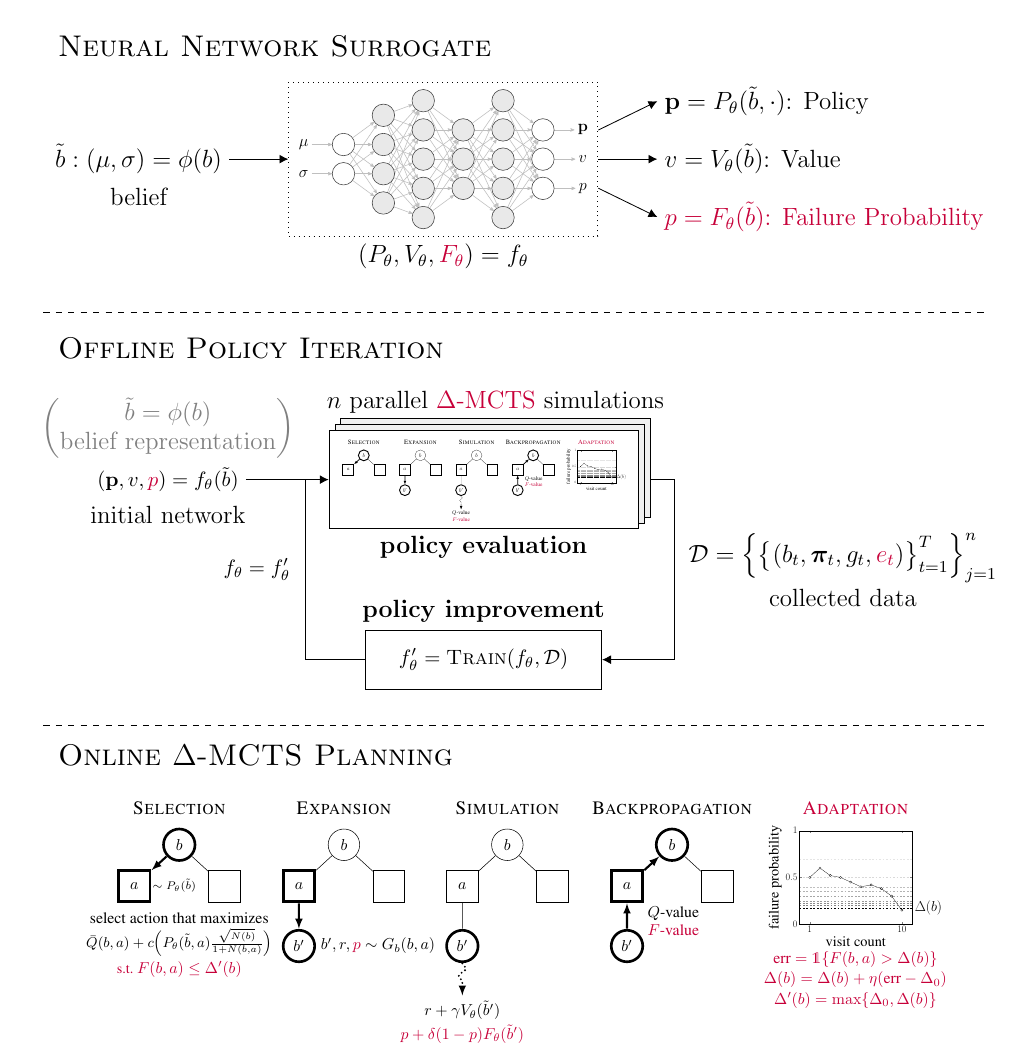}
    }
    \caption{Elements of \textit{ConstrainedZero} for CC-POMDP planning.}
    \label{fig:cbz-alg}
\end{figure}

\section{Introduction}
When developing safety-critical agents to make sequential decisions in uncertain environments, planning and reinforcement learning algorithms often formulate the problem as a partially observable Markov decision process (POMDP) with the objective of maximizing a scalar-valued reward function \cite{dmbook}.
POMDP solution methods find a policies that maximizes this reward.
To ensure adequate safety, the scalar reward is tuned to balance the goals of the agent while penalizing undesired behavior or failures.
Recently, chance-constrained POMDPs (CC-POMDPs) have been used to frame the safe planning problem by separating the reward function into a constrained problem \cite{santana2016rao}.
The objective of CC-POMDPs is to maximize the goal rewards while satisfying the safety constraints.

To solve CC-POMDPs, online algorithms such as RAO$^*$ use heuristic forward search to find policies that maximize the reward and estimate the risk of constraint violation \cite{santana2016rao}.
RAO$^*$ plans over the reachable belief space for discrete state, action, and observation CC-POMDPs.
The iterative RAO$^*$ (iRAO$^*$) extends the heuristic search algorithm to multi-agent settings and handles continuous states and actions through Gaussian process regression and probabilistic flow tubes \cite{huang2018hybrid}.
\citet{lauri2022partially} highlight the limitations of such chance-constrained POMDP algorithms and the need for scalable approaches to solve large-scale, long-horizon CC-POMDPs in practice. 

To address scalability and applicability to continuous state and observation spaces, we introduce the \textit{ConstrainedZero} policy iteration algorithm that combines offline neural network training of the value function, the action-selection policy, and the failure probability predictor with online Monte Carlo tree search (MCTS) to improve the policy through planning.
ConstrainedZero is a direct extension to the POMDP belief-state planning algorithm BetaZero \cite{moss2023betazero} and the family of AlphaZero algorithms \cite{silver2018general}, with extensions shown in red in \cref{fig:cbz-alg}.
Along with an open-source implementation,\footnote{\url{https://github.com/sisl/BetaZero.jl/tree/safety}} our main contributions are threefold:
\begin{enumerate}
    \item We introduce $\Delta$-MCTS, an anytime algorithm for MDPs (applied to belief-state MDPs) that estimates failure probabilities along with $Q$-values and adjusts the failure probability threshold using adaptive conformal inference \cite{gibbs2021adaptive}. $\Delta$-MCTS selects actions by maximizing the $Q$-value while satisfying that the failure probability constraint is below the adapted threshold using the introduced CC-PUCT criterion.
    \item We introduce ConstrainedZero, a policy iteration algorithm that extends BetaZero for CC-POMDPs. ConstrainedZero includes an additional network head that estimates the failure probability given a belief and uses $\Delta$-MCTS with the neural network surrogate to prioritize promising safe actions, replacing expensive rollouts or domain-specific heuristics.
    Framing the problem as a CC-POMDP means a target safety level can be specified instead of balancing penalties in the reward function.
    \item We empirically evaluate ConstrainedZero and $\Delta$-MCTS on three challenging safety-critical benchmark CC-POMDPs: a long-horizon localization task (LightDark \cite{platt2010belief}), an aircraft collision avoidance system (modeled after ACAS X \cite{kochenderfer2012next}), and a CO$_2$ storage agent \cite{corso2022pomdp}.
\end{enumerate}

\section{Problem Formulation}
This section formulates the safe planning problem as a belief-state CC-MDP.
Background is also provided on Monte Carlo tree search and the BetaZero policy iteration algorithm.

\paragraph{POMDPs and belief-state MDPs.}
The partially observable Markov decision process (POMDP) is a framework for sequential decision making problems where the agent has uncertainty over their state in the environment \cite{dmbook}.
The POMDP is a 7-tuple $\langle \mathcal{S}, \mathcal{A}, \mathcal{O}, T, R, O, \gamma \rangle$ consisting of a state space $\mathcal{S}$,
an action space $\mathcal{A}$,
an observation space $\mathcal{O}$,
a transition model $T$,
a reward model $R$,
an observation model $O$,
and a discount factor $\gamma \in [0, 1]$.
When solving POMDPs, the objective is to find a policy $\pi(b)$ given a belief $b$ over the unobserved state and return an action $a \in \mathcal{A}$ that maximizes the \textit{value} of the belief, which is the expected discounted sum of rewards (i.e., the expected discounted \textit{returns}) when continuing to follow the policy $\pi$:
\begin{equation}
    \pi(b_0) = \argmax_{a \in \mathcal{A}} \ \mathbb{E}_\pi\!\left[\sum_{t=0}^\infty \gamma^t R_b(b_t, a_t) \mid b_0 \right]
\end{equation}
where $b_0$ is the initial belief, often using the initial state distribution, and the belief-based reward is defined as
\begin{equation}
R_b(b,a) = \int_{s \in \mathrlap{\mathcal{S}}} b(s)R(s,a) \diff s. \label{eq:belief_reward}
\end{equation}

Every POMDP can be cast as an MDP by simply treating the belief as the state.
In doing so, one can construct a belief-state MDP (BMDP) with the belief space $\mathcal{B}$ of the original POMDP as the MDP state space, while using the same action space $\mathcal{A}$, the belief-based reward model $R_b$ from \cref{eq:belief_reward}, and a transition function $b' \sim T_b(\cdot \mid b, a)$ that takes the current belief $b$ and action $a$ and returns a stochastic updated belief $b'$.
The belief transition function first samples a hidden state $s \sim b(\cdot)$ and transitions that state through the POMDP transition function $s' \sim T(\cdot \mid s, a)$.
Then an observation is sampled from the observation model $o \sim O(\cdot \mid a, s')$ and finally the belief is updated to get the posterior
\begin{equation}
    b'(s') \propto O(o \mid a, s') \int_{s \mathrlap{\in \mathcal{S}}} T(s' \mid s, a)b(s) \diff s. \label{eq:belief_update}
\end{equation}
The belief update may be done exactly as in \cref{eq:belief_update} or using approximations such as a Kalman filter \cite{wan2000unscented} or particle filter \cite{thrun2005probabilistic}.

The belief-state MDP tuple of $\langle \mathcal{B}, \mathcal{A}, T_b, R_b, \gamma \rangle$ can also be defined using a generative model $(b', r) \sim G_b(b, a)$ instead of an explicit belief transition model $T_b$ and belief reward model $R_b$.
The underlying POMDP can also use a generative model $(s', r, o) \sim G(s, a)$.
Our work uses the generative POMDP $\langle \mathcal{S}, \mathcal{A}, \mathcal{O}, G, \gamma \rangle$ and the generative BMDP $\langle \mathcal{B}, \mathcal{A}, G_b, \gamma \rangle$.

\paragraph{Chance-constrained planning.}
When dealing with safety-critical sequential decision making problems, separating safety constraints from the objective allows for solvers to target an adequate level of safety while simultaneously maximizing rewards.
This is in contrast to designing a single reward function to balance the rewards from the goals and penalties from violating safety. 
The chance-constrained POMDP (CC-POMDP) defines a failure set $\mathcal{F}$ that includes all state-action pairs $(s, a) \in \mathcal{S} \times \mathcal{A}$ that fail and a bound $\Delta \in [0, 1]$ on the probability, or chance, of a failure event occurring.
Chance constraints are intuitive for users to define as they translate to the target failure probability of the agent, which is often the requirement for systems in industries such as aviation \cite{busch1985methodology} and finance \cite{flannery1989capital}.
The objective when solving CC-POMDPs is to maximize the value function while ensuring that the failure probability, or the chance constraint, is below the target threshold $\Delta$:
{\small
\begin{alignat}{1}
    \operatornamewithlimits{maximize}_\pi \,\, V^\pi(b_0) &= \mathbb{E}_\pi \Bigg[ \sum_{t=0}^\infty \gamma^t R_b(b_t, a_t) \mid b_0 \Bigg] \\
    \operatorname{subject~to} \,\, F^\pi(b_0) &= \mathbb{P}_\pi \!\left[ \betterbigvee_{t=0}^\infty \Big( (s_t, a_t) \in \mathcal{F} \Big) \mid b_0 \right] \le \Delta
\end{alignat}
}%
The failure probability $F^\pi(b_t)$ is often called the \textit{execution risk} of the policy $\pi$ computed from the belief $b_t$.

Therefore, the CC-POMDP is defined as the tuple $\langle \mathcal{S}, \mathcal{A}, \mathcal{O}, \mathcal{F}, T, R, O, \gamma, \Delta \rangle$ which may also use a generative model $G$ to replace $\langle T,R,O \rangle$.
Our work casts the chance-constrained POMDP to a chance-constrained belief-MDP (CC-BMDP).
The CC-BMDP tuple $\langle \mathcal{B}, \mathcal{A}, F_b, T_b, R_b, \gamma, \Delta \rangle$ extends BMDPs with an immediate failure probability function $F_b: \mathcal{B} \times \mathcal{A} \to [0, 1]$ and a failure probability threshold~$\Delta$.
The immediate failure probability is computed as
\begin{equation}
    F_b(b, a) = \int_{s \mathrlap{\in \mathcal{S}}} b(s) \mathds{1}\big\{(s,a) \in \mathcal{F}\big\} \diff s
\end{equation}
using the failure set $\mathcal{F}$.
The CC-BMDP can also be defined with a generative model $(b', r, p) \sim G_b(b,a)$ that also returns the failure probability $p = F_b(b,a)$ with notation overloading of $G_b$, resulting in the tuple $\langle \mathcal{B}, \mathcal{A}, G_b, \gamma, \Delta \rangle$.

\paragraph{Monte Carlo tree search.}
The best-first search algorithm, Monte Carlo tree search (MCTS), is designed to solve MDPs \cite{coulom2007efficient} and has been applied to solve POMDPs cast as belief-state MDPs \cite{sunberg2018online,fischer2020information,moss2023betazero}.
MCTS is an online algorithm that determines the best action to take from the current state $s$ (or belief state $b$).
Starting from the root state, MCTS iteratively simulates the following four steps to build out a tree of possible reachable futures to a depth $d$:
\begin{enumerate}
    \item \textbf{Selection.}\quad 
    An action is selected from the existing children of the current state node or sampled from the action space.
    The selection process balances exploration and exploitation.
    Metrics such as UCT \cite{kocsis2006bandit} or PUCT \cite{silver2017mastering} have been used in the literature to select the action to take. 
    \item \textbf{Expansion.}\quad 
    Once an action is selected, it is executed from the current state node to expand the tree.
    For stochastic state transitions, methods like progressive widening \cite{couetoux2011continuous} or state abstraction refinement \cite{sokota2021monte} can be used to control when to execute the action or take an existing tree path.
    \item \textbf{Simulation.}\quad 
    From the expanded state, the tree is recursively built from this new root node.
    Simulation returns an estimate of the value of the expanded state.
    The value estimate could use a rollout policy \cite{silver2016mastering} (which may be expensive for BMDPs), or use function approximators such as neural networks \cite{silver2017mastering,fischer2022guiding,moss2023betazero}.
    \item \textbf{Backpropagation.}\quad 
    Finally, the value estimate is combined with the immediate reward to get the $Q$-value.
    This $Q$-value is assigned to the parent state-action node as a running mean.
    This process backpropagates the signal up the tree path that led to that node. 
\end{enumerate}
After a prescribed number of iterations, or at anytime as determined by a compute-time constraint, MCTS will select the best action from the children of the root node, often using the $Q$-values or visit counts \cite{browne2012survey}.

\citet{brazdil2020reinforcement} introduce an algorithm for CC-MDPs that uses UCT with a table-based value and risk predictor, and a linear program to compute an action distribution that satisfies an adaptive constraint.
\citet{ayton2018vulcan} introduce an MCTS algorithm for CC-MDPs that selects actions that satisfy a local chance constraint over state histories and prune branches that violate the constraint.

MCTS algorithms have also been developed for cost-constrained POMDPs (C-POMDPs) \cite{isom2008piecewise}, where a discounted cost is minimized.
Algorithms such as C-POMCP \cite{lee2018monte}, C-MCTS \cite{parthasarathy2023cmcts}, C-POMCPOW, C-PTF-DPW, and C-POMCP-DPW \cite{jamgochian2023online} estimate the cost function during search using rollouts, while our approach uses \textit{chance} constraints and estimates constraint violation \textit{probabilities} using neural network surrogates.

\paragraph{BetaZero.}
The BetaZero belief-state planning algorithm extends AlphaZero to long-horizon POMDPs \cite{moss2023betazero}.
BetaZero combines MCTS planning in belief space with neural network surrogates of the value function and action-selection policy.
The neural network takes as input the belief, which is converted to an input representation $\tilde{b} = \phi(b) = (\mu, \sigma)$ of summary statistics, and estimates the value $v = V_\theta(\tilde{b})$ and policy vector $\mathbf{p} = P_\theta(\tilde{b})$ as two heads of the network $(\mathbf{p}, v) = f_\theta(\tilde{b})$.
BetaZero improves the neural network through policy iteration by iterating the following:
\begin{enumerate}
    \item \textbf{Policy evaluation}: Execute parallel MCTS episodes using the current network to collect training data for policy imitation learning and value regression.
    \item \textbf{Policy improvement}: Use the recent MCTS data over a specified window to re-train the neural network.
\end{enumerate}
Learning from experience through policy iteration reduces the required MCTS depth of search (by estimating the value of future belief states) and breadth of search (by selecting actions prioritized by the policy head of the network).
The neural network surrogate also acts as a learned replacement for domain-specific heuristics of the value function and policy.

In BetaZero, the root node action selection used during MCTS, which is also the trained policy vector target, is a combination of the observed $Q$-values and visit counts $N$ seen during search.
This combination incorporates all available information in the tree, given that belief-state planning is often limited due to the expensive belief updates that occur every state transition.
The root node action is selected according to the $Q$-weighted policy vector:
{\small\begin{equation}
    \bpi_\text{tree}(b, a) \propto \Bigg( \bigg(\frac{\exp Q(b,a)}{\sum_{a'} \exp Q(b,a')}\bigg) \bigg(\frac{N(b,a)}{\sum_{a'} N(b,a')}\bigg) \Bigg)^{1/\tau} \label{eq:policy_q_weight}
\end{equation}}
where $\tau$ controls the sampling temperature (used during policy iteration) and returns the argmax when $\tau \to 0$ (used during final evaluation).
Intuitively, the $Q$-weighted policy vector uses information from what was \textit{found} during search ($Q$-values) and what the search \textit{focused on} (visit counts). 
\citet{moss2023betazero} show that the combination leads to the highest return on various benchmark POMDPs.

\begin{figure}[t!]
    \centering
    \input{algorithms/cbz}
\end{figure}

\section{Approach}
ConstrainedZero follows the BetaZero policy iteration steps of \textit{policy evaluation} and \textit{policy improvement} while also collecting failure event indicators to train the failure probability network head, shown in red in \cref{alg:cbz}.
During policy evaluation, $n$ parallel $\Delta$-MCTS executions are run and a data set $\mathcal{D}$ is collected.
The data set $\mathcal{D} = \big\{\{b_t, \bpi_t, g_t, e_t\}_{t=1}^T \big\}_{j=1}^n$ is a tuple of the belief at episode time step $t$ denoted $b_t$, the tree policy $\bpi_t$, the return $g_t = \sum_{i=t}^T \gamma^{(i-t)}r_i$ based on the observed reward $r_i$ and discount factor $\gamma$, and the failure event indicator $e_t$, where $g_t$ and $e_t$ are computed at the end of the trajectory for all time $t \le T$.
The failure event is computed as the disjunction of all state and action pairs of the CC-POMDP in the execution trajectory to ensure that if a trajectory failed at some point the full trajectory is marked as a failure:
\begin{equation}
    e_t = \mathds{1} \left\{ \betterbigvee_{i=t}^T \Big( (s_i, a_i) \in \mathcal{F} \Big) \right\}
\end{equation}
where $\mathds{1}\{E\}$ is the indicator function that returns $1$ when event $E$ is true and $0$ otherwise.

During policy improvement, the neural network is trained to minimize the MSE or MAE loss $\mathcal{L}_{V_\theta}(g_t, v_t)$ to regress the value function $v_t = V_\theta(\tilde{b}_t)$, minimize the cross-entropy loss $\mathcal{L}_{P_\theta}(\bpi_t, \mathbf{p}_t)$ to imitate the tree policy $\mathbf{p}_t = P_\theta(\tilde{b}_t)$, and additionally minimize the binary cross-entropy loss $\mathcal{L}_{F_\theta}(e_t, p_t)$ to regress the failure probability function $p_t = F_\theta(\tilde{b}_t)$, with added regularization using the $L_2$-norm of the weights $\theta$:
\begin{gather*}
    \mathcal{L}_{V_\theta}(g_t, v_t) = (g_t - v_t)^2 \text{ or } |g_t - v_t|\\
    \mathcal{L}_{P_\theta}(\bpi_t, \mathbf{p}_t) = -\bpi_t^\top \log \mathbf{p}_t\\
    \mathcal{L}_{F_\theta}(e_t, p_t) = -e_t \log p_t - (1-e_t)\log(1-p_t)\\
    \ell_\text{CZ} = \mathcal{L}_{V_\theta}(g_t, v_t) + \mathcal{L}_{P_\theta}(\bpi_t, \mathbf{p}_t) + \mathcal{L}_{F_\theta}(e_t, p_t) + \lambda\norm{\theta}^2
\end{gather*}
The failure probability head of the neural network includes a final sigmoid layer to ensure the output can be interpreted as a probability in the range $[0,1]$.

\begin{figure*}[t!]
    \centering
    \includegraphics[width=0.895\textwidth]{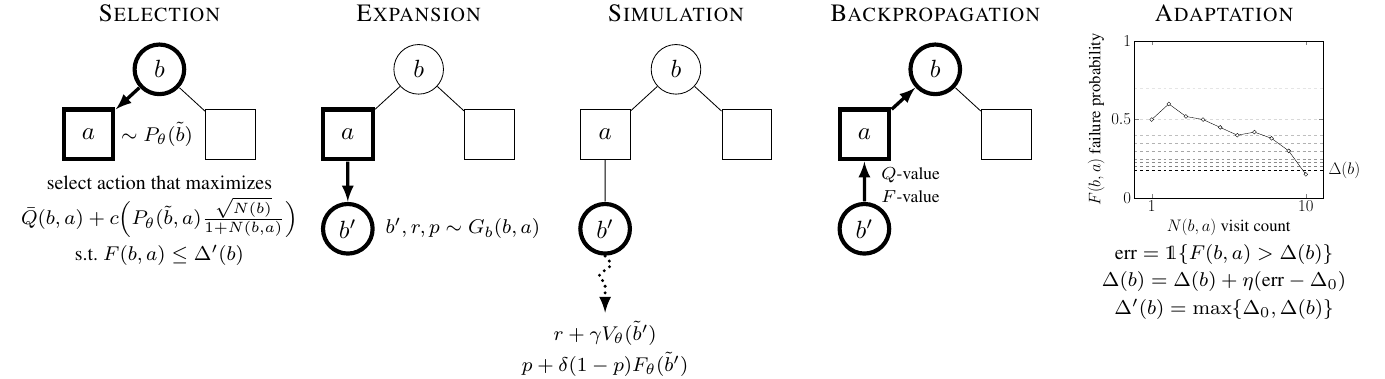}
    \caption{\textit{ConstrainedZero} online Monte Carlo tree search with failure threshold adaptation ($\Delta$-MCTS).}
    \label{fig:cbz-mcts}
\end{figure*}

\begin{figure}[b!]
    \centering
    \input{algorithms/cbz-mcts}
\end{figure}

\begin{figure}[b!]
    \centering
    \input{algorithms/cbz-utils}
\end{figure}

\begin{figure*}[b!]
\begin{floatrow}
    \ffigbox[0.28\textwidth]{%
        \centering
        \includegraphics[width=0.28\textwidth]{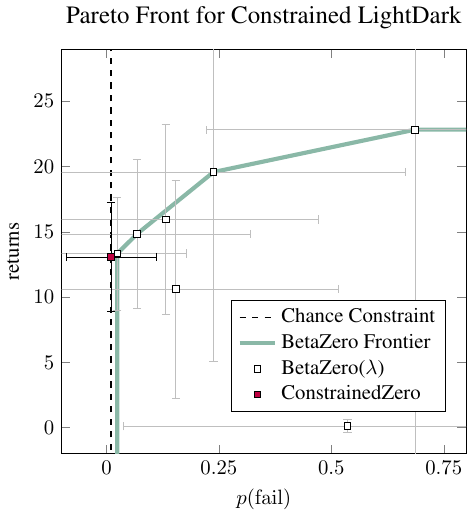}
    }{%
    \caption{BetaZero($\lambda$) comparison.}\label{fig:pareto}
    }
    \capbtabbox[0.66\textwidth]{%
        \centering
        \small
        \begin{threeparttable}
            \begin{adjustbox}{max width=0.66\textwidth}
            \begin{tabular}{@{}lrrrrrrrr@{}}
                \arrayrulecolor{black} 
                \toprule
                  &  \multicolumn{2}{c}{\textbf{LightDark}}  &  &  \multicolumn{2}{c}{\textbf{Collision Avoidance}}  &  &  \multicolumn{2}{c}{\textbf{Spillpoint CCS}} \\
                  &  \multicolumn{2}{c}{\footnotesize$\Delta_0 = 0.01$}  &  &  \multicolumn{2}{c}{\footnotesize$\Delta_0 = 0.01$}  &  &  \multicolumn{2}{c}{\footnotesize{$\Delta_0 = 0.05$}} \\
                \arrayrulecolor{lightgray}
                \cmidrule{2-9}
                \arrayrulecolor{black} 
                                 &  $p(\text{fail}) \downarrow$  &  returns $\uparrow$  &  &  $p(\text{fail}) \downarrow$  &  returns $\uparrow$  &  &  $p(\text{fail}) \downarrow$  &  returns $\uparrow$ \\
                \midrule
                ConstrainedZero  &  $\mathBF{0.01_{\bpm 0.01}}$  &  $\mathBF{13.07_{\bpm 0.42}}$  &  &  $\mathBF{0.00_{\bpm 0.00}}$  &  $\mathBF{-0.74_{\bpm 0.03}}$  &  &  $\mathBF{0.05_{\pm 0.02}}$  &  $\mathBF{2.62_{\pm 0.12}}$  \\
                \arrayrulecolor{white}\midrule
                No Adaptation$^*$  &  $0.66_{\pm 0.05}$  &  $27.47_{\pm 3.90}$  &  &  $0.03_{\pm 0.02}$  &  $-1.00_{\pm 0.00}$  &  &  $0.69_{\pm 0.04}$  &  $6.18_{\pm 0.36}$  \\
                \arrayrulecolor{white}\midrule
                $\Delta$-MCTS (no $f_\theta$)$^\dagger$  &  $0.01_{\pm 0.01}$  &  $1.86_{\pm 0.20}$  &  &  $0.32_{\pm 0.05}$  &  $0.00_{\pm 0.00}$  &  &  $1.00_{\pm 0.00}$  &  $6.87_{\pm 0.50}$  \\
                \arrayrulecolor{grays1}\midrule
                Raw Policy $P_\theta$  &  $0.01_{\pm 0.01}$  &  $12.88_{\pm 0.46}$  &  &  $0.00_{\pm 0.00}$  &  $-0.86_{\pm 0.02}$  &  &  $0.06_{\pm 0.02}$  &  $2.45_{\pm 0.11}$  \\
                \arrayrulecolor{white}\midrule
                Raw Value$^\ddagger$ $V_\theta$  &  $0.72_{\pm 0.05}$  &  $28.00_{\pm 4.51}$  &  &  $0.16_{\pm 0.04}$  &  $-0.20_{\pm 0.04}$  &  &  $0.38_{\pm 0.05}$  &  $4.27_{\pm 0.30}$  \\
                \midrule
                Raw Failure$^\ddagger$ $F_\theta$  &  $0.80_{\pm 0.04}$  &  $0.05_{\pm 0.04}$  &  &  $0.00_{\pm 0.00}$  &  $-1.62_{\pm 0.08}$  &  &  $0.00_{\pm 0.00}$  &  $0.00_{\pm 0.00}$  \\
                \arrayrulecolor{black}
                \bottomrule
            \end{tabular}
            \end{adjustbox}
            \begin{scriptsize}
                \begin{tablenotes}
                    \linespread{1.05}\selectfont 
                    \item[\!\!] {All results report the mean $\pm$ standard error over \num{100} seeds, evaluated using the argmax of \protect\cref{eq:policy_q_weight}, i.e., $\tau \to 0$.}
                    \item[*] {Trained with the same parameters as ConstrainedZero without adaptation, i.e., only a hard constraint on $\Delta_0$.}
                    \item[$\dagger$] {$\Delta$-MCTS without the neural network for the value or failure probability and a random policy for CC-PUCT.}
                    \item[$\ddagger$] {One-step look-ahead over all actions using only the value or failure probability network head with \num{5} obs. per action.}
                \end{tablenotes}
            \end{scriptsize}
        \end{threeparttable}
    }{%
        \caption{ConstrainedZero results. Bold indicates the best results within the $\Delta_0$ threshold.}
        \label{tab:results}    
    }
\end{floatrow}
\end{figure*}

\subsection{Adaptive safety constraints in \texorpdfstring{$\Delta$}{Δ}-MCTS}
When using online MCTS for CC-BMDP planning, two considerations have to be addressed: 1) how to estimate the observed failure probability in the tree search, and 2) how to select actions constrained by this failure probability.

At each node for the belief-state and action $(b,a)$, the immediate failure probability $p$ is computed using the generative model (or by calling $p = F_b(b,a)$ directly). 
An estimate of the future failure probability $p'$ can be computed using rollouts, which may be expensive for belief-state planning, thus we use the trained neural network head for failure probability estimation $p' = F_\theta(\tilde{b}')$.
Similar to the $Q$-value, we must compute the full failure probability of the trajectory from the immediate time step to the horizon, termed the $F$-value.
Let $E$ be the immediate failure event from belief $b$ when taking action $a$ at time $t$, and let $E'$ be the event of failing in the future (from $t+1$ to the horizon $T$).
The probability of failing between the current time $t$ and the horizon $T$ becomes:
\begin{align}
    P(E_{t:T}) &= P(E) + P(E') - P(E \cap E')\\
               &= P(E) + P(E') - P(E')P(E \mid E')\\
               &= P(E) + P(E') - P(E')P(E) \label{eq:p_indep}\\
               &= p + p' - pp'\\
               &= p + (1-p)p'
\end{align}
assuming independence in \cref{eq:p_indep}.
A discount $\delta$ is applied to control the influence of the future failure probability:
\begin{equation}
    p = p + \delta(1-p)p' \label{eq:p_fail}
\end{equation}
Unlike \citet{carpin2022solving}, who backup $F$-values based on the best-case, we backpropagate the $F$-values up the tree similar to $Q$-values (alg. \ref{alg:cbz-mcts}, line \ref{line:f_value}):
\begin{equation}
    F(b,a) = F(b,a) + \frac{p - F(b,a)}{N(b,a)} \label{eq:f_value}
\end{equation}
which is a running mean estimate where $F(b,a)$ is initialized using the initialization function $F_0(b,a)$ (noting the $F_0$ subscript: which could either be zero, the immediate failure probability $F_b(b,a)$, or the bootstrapped value by taking action $a$ to get a new belief $b'$ and computing \cref{eq:p_fail} based on the $p' = F_\theta(\tilde{b}')$ estimate).
Note, the number of times the node $(b,a)$ is visited is indicated as the visit count $N(b,a)$.

Using the estimate $F(b,a)$, a simple way to select actions that do not violate the safety constraint set by $\Delta$ would be to use the PUCT algorithm \cite{silver2018general}
{\small\begin{align}
    \pi_\text{explore}(b) = \argmax_{a \in A(b)}& \ \bar{Q}(b, a) + c \Big(P_\theta(\tilde{b},a) \textstyle\frac{\sqrt{N(b)}}{1 + N(b,a)} \Big)
\end{align}}
with a hard constraint on safety of only choosing actions such that $F(b,a) \le \Delta$ is satisfied.
PUCT exploits nodes based on their observed $Q$-values and explores nodes based on their visit counts weighted by the action-selection policy $P_\theta$ to explore promising actions.\footnote{The belief node visit count is $N(b) = \sum_{a'} N(b,a')$ for children ${a' \in A(b)}$. Following \citet{schrittwieser2020mastering}, we normalize the $Q$-values between zero and one, denoted $\bar{Q}$, to avoid problem-specific heuristics when selecting an exploration constant $c$:
{\small
\begin{gather}
    \bar{Q}(b,a) = \frac{Q(b, a) - \min_{(b',a') \in \mathcal{T}} Q(b', a')}{\max_{(b',a') \in \mathcal{T}} Q(b', a') - \min_{(b',a') \in \mathcal{T}} Q(b', a')}
\end{gather}%
}%
}

However, if the failure probability threshold $\Delta$ is too conservative, the action-selection process may fail to find \textit{any} action that satisfies the constraint.
Therefore, $\Delta$-MCTS tracks an estimate of the threshold $\Delta(b)$ for each belief node and updates it using \textit{adaptive conformal inference} (ACI) \cite{gibbs2021adaptive}.
ACI is a statistical method that provides valid prediction intervals without assumptions on how the time-series data was generated.
The adaptive threshold is initialized to the target tolerance $\Delta(b) = \Delta_0$ where $\Delta_0 = \Delta$ from the CC-BMDP.
Each time the $F$-value is updated (either by \cref{eq:f_value} or initialization), the \textproc{Adaptation} procedure is called to update the current acceptable safety threshold.

In adaptation, the error term of $\text{err} = \mathds{1}\{F(b,a) > \Delta(b)\}$ indicates when to widen or restrict the estimated threshold $\Delta(b)$ based on whether the failure probability estimate of the most recently explored belief-action node $F(b,a)$ is above or below the current threshold.
The estimated threshold is updated according to 
\begin{equation}
    \Delta(b) = \Delta(b) + \eta(\text{err} - \Delta_0) \label{eq:aci_update}
\end{equation}
which will widen the threshold if the observed failure probability is outside the threshold (i.e., if the error is one), and will tighten the threshold otherwise:
\begin{equation}
    \Delta(b) = \begin{cases}
        \Delta(b) + \eta(1-\Delta_0) & \text{if } F(b,a) > \Delta(b)\\
        \Delta(b) - \eta\Delta_0 & \text{if } F(b,a) \leq \Delta(b)
    \end{cases} \label{eq:err}
\end{equation}
Intuitively, the update adjusts the threshold of acceptable failure probability $\Delta(b)$ based on recent experience.
If the failure probability $F(b,a)$ for a recent action is higher than the current threshold $\Delta(b)$, this indicates a higher risk than expected.
Thus, the threshold is increased by $\eta(1-\Delta_0)$ for $\eta > 0$ to allow for more risk in future actions.
Otherwise, if $F(b,a)$ is lower than the threshold, this means actions are safer than expected and the threshold is decreased by $\eta\Delta_0$ (favoring a more reactive increase than decrease of the threshold).
Notably, \citet{gibbs2021adaptive} prove that $\Delta(b)$ converges exactly to the desired target over time.
The appendix details an analysis of $\eta$ and our simplification of ACI for threshold adaptation and how it is a reformulation of quantile coverage.

We clip the final threshold to the lower and upper bounds of the observed failure probability for a given belief $b$ to restrict the change in $\Delta(b)$ and, more importantly, to guarantee that at least one action is available for selection:
\begin{equation}
    \Delta(b) = \operatorname{clip}\!\big( \Delta(b) + \eta(\text{err} - \Delta_0),\, l(b),\, u(b) \big)
\end{equation}
with the lower and upper bounds of $l(b) = \min_{a'} F(b, a')$ and $u(b) = \max_{a'} F(b, a')$ for $a'$ in children nodes $A(b)$.

The resulting criterion selects actions that satisfy the adaptive constraint of $F(b,a) \le \Delta'(b)$ where the selection threshold $\Delta'(b) = \max \{\Delta_0, \Delta(b)\}$ upper bounds the failure probability.
Together, the $\Delta$-MCTS exploration policy becomes:
{\small\begin{align}
    \pi_\text{explore}(b) = \argmax_{a \in A(b)}& \ \bar{Q}(b, a) + c \Big(P_\theta(\tilde{b},a) \textstyle\frac{\sqrt{N(b)}}{1 + N(b,a)} \Big)\\
                             \operatorname{s.t.}& \ F(b,a) \le \Delta'(b) \label{eq:constraint}
\end{align}}
termed the \textit{chance-constrained PUCT} criterion (CC-PUCT).
The constraint in \cref{eq:constraint} is also used in root node action selection (line \ref{line:root_constraint}, alg. \ref{alg:cbz-mcts}).
In practice, $\pi_\text{explore}$ is computed using the indicator $\mathds{1}$, returning the action $a \in A(b)$ that maximizes:
\begin{equation*}
\mathds{1}\big\{F(b,a) \le \Delta'(b) \big\} \Big(\bar{Q}(b, a) + c \Big(P_\theta(\tilde{b},a) \textstyle\frac{\sqrt{N(b)}}{1 + N(b,a)} \Big) \Big)
\end{equation*}

The benefit of CC-PUCT is that when our explored samples satisfy the constraint $\Delta'(b)$ (defined over the belief rather than both belief and action) we may explore new actions from this belief which are both safe and have the potential for higher reward.
The key idea is that actions are chosen based on the balance between safety and utility; ensuring that we do not over-prioritize safety at the expense of potential rewards, while not exploiting rewards without regarding the risk.

The five stages of $\Delta$-MCTS are shown in algorithms \ref{alg:cbz-mcts}--\ref{alg:cbz-utils}: \textit{selection}, \textit{expansion}, \textit{simulation}, \textit{backpropagation}, and \textit{adaptation}, with extensions to BetaZero shown in red.

\begin{figure*}[ht!]
    \centering
    \begin{subfigure}[b]{0.245\textwidth}
        \resizebox{\textwidth}{!}{%
            \includegraphics{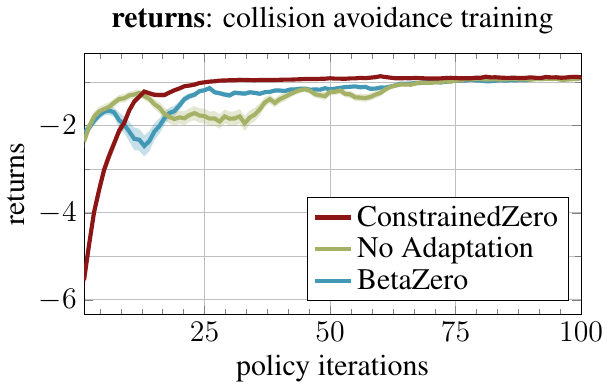}
        }
        \caption{Returns from policy iteration.}\label{fig:cas_results_returns}
    \end{subfigure}
    \hfill
    \begin{subfigure}[b]{0.245\textwidth}
        \resizebox{\textwidth}{!}{%
            \includegraphics{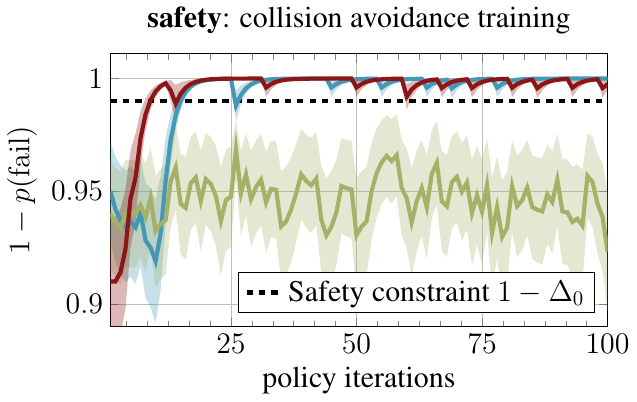}
        }
        \caption{Safety from policy iteration.}\label{fig:cas_results_safety}
    \end{subfigure}
    \hfill
    \begin{subfigure}[b]{0.245\textwidth}
        \resizebox{\textwidth}{!}{%
            \includesvg[pretex=\LARGE]{figures/results/ijcai_cas_pfail_head.svg}
        }
        \caption{Higher $p(\text{fail})$ near an NMAC.}\label{fig:cas_results_pfail}
    \end{subfigure}
    \hfill
    \begin{subfigure}[b]{0.245\textwidth}
        \resizebox{\textwidth}{!}{%
            \includesvg[pretex=\LARGE]{figures/results/ijcai_cas_trajs.svg}
        }
        \caption{Climb (blue), descend (green).}\label{fig:cas_results_trajs}
    \end{subfigure}
    \caption{Results for the collision avoidance CC-POMDP. \Cref{fig:cas_results_trajs} matches the ``notch'' behavior from \protect\citet{kochenderfer2012next}.}\label{fig:cas_results}
\end{figure*}

\section{Experiments}
For a fair comparison, ConstrainedZero was evaluated against BetaZero using the same network and MCTS parameters.
BetaZero uses a scalarized reward function to penalize failures, while ConstrainedZero omits the penalty and plans using the adaptive safety constraint instead.
The BetaZero reward takes the form $\bar{R}_b(b,a) = R_b(b,a) - \lambda C(b,a)$ with a cost $C$ scaled by $\lambda$.
Three safety-critical CC-POMDPs were evaluated.

\paragraph{LightDark localization.}
The \textit{LightDark} POMDP is a standard benchmark localization task where the agent can move either up or down by one to localize in a one-dimensional space with the goal of stopping at the origin \cite{platt2010belief}.
The agent receives noisy observations of its $y$-state position as a function of the distance to the light region at ${y=10}$.
In the POMDP used by BetaZero, the agent receives a reward of $100$ for stopping at $\pm 1$ of the origin, and a penalty of $-100$ for stopping outside of the origin.
In the CC-POMDP, the agent only receives positive reward for executing stop at the origin and an indication of failure occurs in the event of the agent stopping outside the origin.
A particle filter is used to update the belief with $n_\text{particles}=500$.

\paragraph{Aircraft collision avoidance.}
The aircraft collision avoidance system (CAS) is modeled after ACAS X, a real-world application of POMDPs \cite{kochenderfer2012next}, where the ownship aircraft avoids a near mid-air collision (NMAC) with an intruding aircraft while minimizing the alert and reversal rates.
The ownship can apply vertical rate changes in $[-5, 0, 5]$\,\si{\meter/\second} to maneuver away from the intruder.
The state variables are the relative altitude $h_\text{rel}$, relative vertical rate $\dot{h}_\text{rel}$, previous action $a_\text{prev}$, and time to collision $\tau$ \cite{dmbook}.
An NMAC occurs when $|h_\text{rel}| \le 50$ and $\tau\,{=}\,0$.
Beliefs are updated with an unscented Kalman filter to track the mean and covariance of the state \cite{wan2000unscented}.
For the POMDP, a penalty of $-1$ is incurred for the first alert or when reversing the action and $-100$ if an NMAC occurs.
No NMAC penalty is used in the CC-POMDP.

\paragraph{Safe carbon storage.}
Carbon capture and storage (CCS) is a promising mitigation of global emissions that captures CO$_2$ and stores it in porous subsurface material \cite{corso2022pomdp}.
A challenge of CCS is safely injecting CO$_2$ into the subsurface while mitigating risk of leakage and earthquakes.
The simplified CCS problem uses spillpoint analysis to model the top surface of the injection site and a sequential importance resampling particle filter is updated with observations at drilling locations. 
The agent can drill, observe, change the injection rate, or stop the project.
A large penalty is received for any leaked CO$_2$, a small penalty for observing, and a reward for trapped CO$_2$.
In the CC-POMDP, no penalty is applied and instead any leakage indicates a failure.

\subsection{Empirical results}
\Cref{fig:pareto} compares ConstrainedZero against BetaZero, where BetaZero uses different values of the penalty $\lambda$.
The penalties were swept between $-10$ and $-1000$ with $-100$ being the standard for the LightDark POMDP (proportional to the goal reward of $100$).
A target safety level of ${\Delta_0 = 0.01}$ was chosen for ConstrainedZero.
ConstrainedZero exceeds the BetaZero Pareto curve and achieves the target level of safety with a failure probability of ${0.01 \pm 0.01}$ computed over $100$ episodes.
Notably, ConstrainedZero also has less variance in the failure probability and the returns.
BetaZero still achieves good performance but at the cost of sweeping the penalty values without explicitly defining a safety threshold to satisfy.

Shown in \cref{tab:results}, an ablation study is conducted for ConstrainedZero.
Most notably, the adaptation procedure is crucial to enable the algorithm to properly balance safety and utility during planning (also shown in \cref{fig:cas_results_returns}--\ref{fig:cas_results_safety}).
When comparing $\Delta$-MCTS without network approximators against ConstrainedZero, it is clear that offline policy iteration allows for better online planning.
Using only the raw policy head $P_\theta$ achieves good performance, which is trained to imitate the tree policy.
However, incorporating additional online planning with the full network yields better results overall, enabling planning over potentially unseen information.
The full ConstrainedZero algorithm consistently achieves the highest return within the satisfied target level of safety $\Delta_0$.

Compared to BetaZero, \cref{fig:cas_results_returns} and \cref{fig:cas_results_safety} highlight that ConstrainedZero satisfies the safety constraint earlier during policy iteration, while simultaneously maximizing returns (shown for the CAS problem).
The policy trained without adaptation learns to maximize returns but fails to satisfy the safety constraint.
This is because without adaptation, the algorithm will attempt to satisfy a fixed constraint, not taking into account the outcomes of its actions.
With adaptation, ConstrainedZero adjusts the constraint in response to feedback from the environment, resulting in the algorithm becoming more capable at optimizing its performance within the bounds of the adaptive constraint.
This demonstrates the importance of adaptation, as a fixed constraint may be too conservative or too risky, leading to suboptimal decision-making.

\section{Conclusions}
This work introduces \textit{ConstrainedZero}, an extension of the BetaZero POMDP planning algorithm to chance-constrained POMDPs.
Along with neural network estimates of the value function and action-selection policy, we include a network head that estimates the failure probability given a belief.
By formulating the safe planning problem as a CC-POMDP, we select a target level of safety to optimize towards, instead of traditional POMDP methods that tune the reward function to balance safety and utility as a multi-objective problem.
We develop an extension to Monte Carlo tree search that includes an \textit{adaptation} stage that adjusts the target level of safety during planning with adaptive conformal inference.
The resulting $\Delta$-MCTS algorithm modifies MCTS for CC-POMDPs and addresses the issue of overfitting to failure predictions.
Additionally, a constrained action-selection criterion (CC-PUCT) was developed to enable planning under constraints.
The full implementation and experiments are open sourced and are part of the Julia package BetaZero.jl.\footnote{\url{https://github.com/sisl/BetaZero.jl/tree/safety}}

\paragraph{Limitations.}
Adapting the safety level when planning with approximations may lead to deviations despite ACI convergence guarantees, but the lack of adaptation also does not guarantee the desired safety.
However, our experiments show that this flexibility helps the algorithm find policies matching the targeted safety.
Similar to BetaZero, ConstrainedZero may require more computing resources than existing POMDP solvers due to neural network training and parallel $\Delta$-MCTS episodes.
However, it is designed for large-scale, uncertain problems in high-dimensional spaces that require long-horizon planning.
We focus on real-world scenarios where transition dynamics, not policy training, are the main challenge, using past experiences to learn an approximately optimal policy offline that is refined online using tree search.

\paragraph{Future work.}
A benefit of $\Delta$-MCTS unexplored in this work is that it also applies directly to safety-critical MDPs,
such as scheduling \cite{soualhia2020dynamic}.
Future work could focus on the application of ConstrainedZero to fully observable MDP settings and extending the algorithm to work over multiple failure modes.
Other future work could improve the data efficiency of ConstrainedZero, extend it to continuous actions similar to algorithms such as A0C \cite{moerland2018a0c}, or apply it to safety-critical robotics tasks such as navigation \cite{sylvie2010planning} and object manipulation \cite{pajarinen2017robotic,pajarinen2022pomdp}.

    \ifdefined\SUBMISSION\else
        \section*{Acknowledgments}
This research is funded by OMV and J.F. thanks the Karlsruhe House of Young Scientists (KHYS) for travel grant funding.

    \fi

    \bibliographystyle{preamble/named}
    \bibliography{references}

    \ifdefined\SUBMISSION\else
        
    \fi
\fi

\end{document}